\documentclass[11pt,a4paper]{article}
\usepackage[hyperref]{emnlp2019}
\usepackage{times}
\usepackage{latexsym}
\usepackage{mwe}
\usepackage{caption}
\usepackage{url}
\usepackage{graphicx} 
\usepackage{subcaption}
\usepackage{enumitem}

\aclfinalcopy 

\usepackage{makecell}
\usepackage{color}

\definecolor{MKS}{rgb}{0.7, 0, 0.1}
\definecolor{AC}{rgb}{0.0, 0.1, 0.7}
\definecolor{EAC}{rgb}{0.0, 0.8, 0.1}
\definecolor{AJ}{rgb}{0.5, 0.5, 0.5}

\title{Sampling Bias in Deep Active Classification: An Empirical Study}

\author{Ameya Prabhu$^1$\thanks{ \hspace{0.1cm} indicates equal contribution}  \hspace{0.05cm}\thanks{ \hspace{0.1cm} Work done at Verisk $\vert$ AI}  , Charles Dognin$^2$\footnotemark[1] , Maneesh Singh$^2$ \\
University of Oxford$^1$ \qquad Verisk $\vert$ AI, Verisk Analytics$^2$ \\
ameya.prabhu@eng.ox.ac.uk, \{charles.dognin, maneesh.singh\}@verisk.com\\}
 
\begin{document}
\maketitle
\begin{abstract}
The exploding cost and time needed for data labeling and model training are bottlenecks for training DNN models on large datasets. Identifying smaller representative data samples with strategies like active learning can help mitigate such bottlenecks. Previous works on active learning in NLP identify the problem of sampling bias in the samples acquired by uncertainty-based querying and develop costly approaches to address it. Using a large empirical study, we demonstrate that active set selection using the posterior entropy of deep models like FastText.zip (FTZ) is robust to sampling biases and to various algorithmic choices (query size and strategies) unlike that suggested by traditional literature. We also show that FTZ based query strategy produces sample sets similar to those from more sophisticated approaches (e.g ensemble networks). Finally, we show the effectiveness of the selected samples by creating tiny high-quality datasets, and utilizing them for fast and cheap training of large models. Based on the above, we propose a simple baseline for deep active text classification that outperforms the state-of-the-art. We expect the presented work to be useful and informative for dataset compression and for problems involving active, semi-supervised or online learning scenarios. Code and models are available at: \href{https://github.com/drimpossible/Sampling-Bias-Active-Learning}{https://github.com/drimpossible/Sampling-Bias-Active-Learning}

\end{abstract}
\section{Introduction}
\label{sec:introduction}

Deep neural networks (DNNs) trained on large datasets provide state-of-the-art results on various NLP problems \cite{devlin2018bert} including text classification \cite{howard2018universal}. However, the cost and time needed to get labeled data and to train models is a serious impediment to creating new and/or better models. This problem can be mitigated by creating smaller representative datasets with active learning which can be used for training DNNs to achieve similar test accuracy as that using the full training dataset . In other words, the smaller sample can be considered a surrogate for the full data. 

However, there is lack of clarity in the active learning literature regarding sampling bias in such surrogate datasets created using active learning \cite{settles2009active}: its dependence on models, functions and parameters used to acquire the sample. Indeed, what constitutes a good sample? In this paper, we perform an empirical investigation using active text classification as the application.
 
Early work in active text classification \cite{lewis1994sequential} suggests that greedy query generation using label uncertainty may lead to efficient representative samples (Nonetheless, the same test accuracy). Subsequent concerns regarding sampling bias has lead to explicit use of expensive diversity measures \cite{brinker2003incorporating, hoi2006large} in acquisition functions or using ensemble approaches \cite{liere1997active, mccallum1998employing} to improve diversity implicitly. 

Deep active learning approaches adapt the discussed framework above to train DNNs on large data. However, it is not clear if the properties of deep approaches mirror those of their shallow counterparts and if the theory and the empirical evidence regarding sampling efficiency and bias translates from shallow to deep models. For example, \cite{sener2018active} and \cite{ducoffe2018adversarial} find that uncertainty based strategies perform no better than random sampling even if ensembles are used and using diversity measures outperform both. On the other hand, \cite{beluch2018power, gissin2019discriminative} find that uncertainty measures computed with ensembles outperform diversity based approaches while \cite{gal2017deep, beluch2018power, siddhant2018deep} find them to outperform uncertainty measures computed using single models. A recent empirical study \cite{siddhant2018deep} investigating active learning in NLP suggests that Bayesian active learning outperforms classical uncertainty sampling across all settings. However, the approaches have been limited to relatively small datasets. 

\subsection{Sampling Bias in Active Classification}
In this paper, we investigate the issues of sampling bias and sample efficiency, the stability of the actively collected query and train sets and the impact of algorithmic factors  - i.e. the setup chosen while training the algorithm, in the context of deep active text classification on large datasets. In particular, we consider two sampling biases: label and distributional bias, three algorithmic factors: initial set selection, query size and query strategy along with two trained models and four acquisition functions on eight large datasets. 

To isolate and evaluate the impact of the above (combinatorial) factors, a large experimental study was necessary. Consequently, we conducted over 2.3K experiments on 8 popular, large, datasets of sizes ranging from 120K-3.6M. Note that the current trend in deep learning is to train large models on very large datasets. However, the aforementioned issues have not yet been investigated in the literature in such a setup. As shown in Table \ref{tab:comparison_datasets}, the datasets used in latest such analysis on active text classification by \cite{siddhant2018deep} are quite small in comparison. The datasets used by us are two orders of magnitude larger, our query samples often being the size of the entire datasets used by previous works, and the presented empirical study is more extensive (20x experiments).
\\

\noindent Our findings are as follows:
 
 (i) We find that utilizing the uncertainty query strategy using a deep model like FastText.zip (FTZ)\footnote{We use FastText.zip (FTZ) to optimize the time and resources needed for this study.} to actively construct a representative sample provides query and train sets with remarkably good sampling properties. 
 
 (ii) We finds that a single deep model (FTZ) used for querying provides a sample set similar to more expensive approaches using ensemble of models. Additionally, the sample set has a large overlap with support vectors of an SVM trained on the entire dataset largely invariant to a variety of algorithmic factors, thus indicating the robustness of the acquired sample set.   
 
(iii) We demonstrate that the actively acquired training datasets can be utilized as small, surrogate training sets with a 5x-40x compression for training large, deep text classification models. In particular, we can train the ULMFiT \cite{howard2018universal} model to state of the art accuracy at 25x-200x speedups.

(iv) Finally, we create a novel, state-of-the-art baseline for active text classification which outperforms recent work \cite{siddhant2018deep}, using Bayesian dropout, utilizing 4x less training data. We also outperform \cite{sener2018active}  at all training data sizes. The latter uses an expensive diversity based query strategy (coreset sampling). 

The rest of the paper is organized as follows: in Section \ref{sec:method}, the experimental methodology and setup are described. Section \ref{sec:results} presents the experimental study on sampling biases as well as the impact of various algorithmic factors. In Section \ref{sec:comparison}, we compare with prior literature in active text classification. Section \ref{sec:implications} presents a downstream use case - fast bootstrapping of the training of very large models like ULMFiT. Finally, we discuss the current literature in light of our work in Section \ref{sec:related} and summarize the conclusions in Section \ref{sec:conclusion}.

\section{Methodology}\label{sec:method}
This section describes the experimental approach and the setup used to empirically investigate the issues of (i) sampling bias and (ii) sampling efficiency in creating small samples to train deep models.

\subsection{Approach}
A labelled training set is incrementally built from a pool of unlabeled data by selecting \& acquiring labels from an oracle in sequential increments. In this, we follow  the standard approach found in the active learning literature. We use the following terminology:

\textbf{Queries \& Query Strategy:} We refer to the (incremental) set of points selected to be labeled and added to the training as the query and the (acquisition) function used to select the samples as the query strategy. 

\textbf{Pool \& Train Sets:} The pool is the unlabeled data from which queries are iteratively selected, labeled and added to the (labeled) train set. 

Let $\mathcal{D}_S = (\mathbf{x}_i, y_i)$ denote a dataset consisting of $\vert \mathcal{S} \vert  = n$ i.i.d samples of data/label pairs, where $\vert . \vert$ denotes the cardinality. Let $\mathcal{S}_0 \subset \mathcal{S}$ denote an initial randomly drawn sample from the initial pool. At each iteration, we train the model on the current train set and use a model-dependent query strategy to acquire new samples from the pool, get them labeled by an oracle and add them to the train set. Thus, a sequence of training sets: $[ \mathcal{S}_1,\mathcal{S}_2\ldots,\mathcal{S}_{b}]$ is created by sampling $b$ queries from the pool set, each of size $K$. The $b$ queries are given by $[ \mathcal{S}_1-\mathcal{S}_0,\mathcal{S}_2-\mathcal{S}_1\ldots,\mathcal{S}_{b}-\mathcal{S}_{b-1}]$. Note that $|\mathcal{S}_i| = (|\mathcal{S}_0| + i \times K)$ and $\mathcal{S}_1 \subset \mathcal{S}_2 \ldots \subset \mathcal{S}_b \subset \mathcal{S}$. 

In this paper, we investigate the efficiency and bias of sample sets $\mathcal{S}_{b}^{1}, \mathcal{S}_{b}^{2}, \ldots, \mathcal{S}_{b}^{t}$ obtained by different query strategies $Q^1, Q^2, \ldots Q^t$. We exclude the randomly acquired initial set and perform comparisons on the actively acquired sample sets defined as $\hat{\mathcal{S}}_{j}^i = (\mathcal{S}_{j}^{i} - \mathcal{S}_0^{i})$.

\subsection{Experimental Setup}

In this section, we share details of the experimental setup, and present and explain the choice of the datasets, models and query strategies used. 

\textbf{Datasets:} We used eight, large, representative datasets widely used for text classification: AG-News (AGN), DBPedia (DBP), Amazon Review Polarity (AMZP), Amazon Review Full (AMZF), Yelp Review Polarity (YRP), Yelp Review Full (YRF), Yahoo Answers (YHA) and Sogou News (SGN). Please refer to Section 4 of \cite{zhang2015character} for details regarding the collection and characteristics of these datasets. Table \ref{tab:comparison_datasets} provides a comparison regarding the choice of datasets, models and number of experiments between our study and \cite{siddhant2018deep} which investigates a variety of NLP tasks including text classification while we focus only on the latter.

\textbf{Models:} We reported two text classification models as representatives of classical and deep learning approaches respectively which were fast to train and also had good performance on text classification:  Multinomial Naive Bayes (MNB) with TF-IDF \cite{wang2012baselines} and FastText.zip (FTZ) \cite{joulin2016fasttext}. The FTZ model provides results competitive with VDCNNs (a 29 layer CNN) \cite{conneau2016very} but with over 15,000$\times$ speedup \cite{joulin2016bag}.  
This allowed us to conduct a thorough empirical study on large datasets. Multinomial Naive Bayes (MNB) with TF-IDF features is a popularly claimed baseline for text classification \cite{wang2012baselines}.


\textbf{Query Strategies:} Uncertainty based query strategies are widely used and well studied in the active learning literature. Those strategies typically use a scoring function on the (softmax) output of a single model. We evaluate the following ones: Least Confidence (LC) and Entropy (Ent). Independently training ensembles of models \cite{lakshminarayanan2017simple} is another principled approach to obtain uncertainties associated with the output estimate.
Then, we tried four query strategies - LC and Ent computed using single and ensemble models and evaluated them against random sampling (chance) as a baseline. For ensembles, we used five FTZ ensembles \cite{lakshminarayanan2017simple}. In contrast, \cite{siddhant2018deep} used  Bayesian ensembles using Dropout, proposed in \cite{gal2017deep}. Please refer to Section \ref{sec:comparison} for a comparison.   

\begin{table}[t]
    \centering
    \resizebox{\columnwidth}{!}{
    \begin{tabular}{|l|c|c|c|c|} \hline
        Paper & \#Exp & Datasets (\#Train) & Models (Full Acc) \\ \hline
        DAL & 120 & \makecell{TREC-QA (6k),\\ MAReview (10.5k)} & \makecell{SVM (89\%), \\CNN (91\%),\\ LSTM (92\%)} \\ \hline
        Ours & 2.3K & \makecell{AGN (120k), SGN (450k),\\ DBP (560k), YRP (560k),\\ YRF (650k), YHA (1400k),\\ AMZP (3600k), AMZF (3000k)} & \makecell{FTZ (97\%), MNB (90\%)}  \\\hline
    \end{tabular}}
    \caption{Comparison of active text classification datasets and models (Acc on Trec-QA) used in \cite{siddhant2018deep} and our work. We use significantly larger datasets (two orders larger), perform 20x more experiments, and use more efficient and accurate models.}
    \label{tab:comparison_datasets}
\end{table}

\textbf{Implementation Details:} We performed 2304 active learning experiments. We obtained our results on three random initial sets and three runs per seed (to account for stochasticity in FTZ) for each of the eight datasets. The query sizes were $0.5\%$ of the dataset for AGN, AMZF, YRF and YHA and $0.25\%$ for SGN, DBP, YRP and AMZP respectively for $b=39$ sequential, active queries. We also experimented with different query sizes keeping the size of the final training data $b \times K$ constant. The default query strategy uses a single model with output Entropy (Ent) unless explicitly stated otherwise. Results in the chance column are obtained using random query strategy.

We used Scikit-Learn \cite{scikitlearn} implementation for MNB and original implementation for FastText.zip (FTZ)  \footnote{\url{https://github.com/facebookresearch/fastText}}. 
We required 3 weeks of running time for all FTZ experiments on a x1.16xlarge AWS instance with Intel Xeon E7-8880 v3 processors and 1TB RAM to obtain results presented in this work. The experiments are deterministic beyond the stochasticity involved in training the FTZ model, random initialization and SGD updates. The entire list of hyperparameters and metrics affecting uncertainty such as calibration error \cite{guo2017calibration} is given in the supplementary material. The experimental logs and models are available on our github link\footnote{\href{https://github.com/drimpossible/Sampling-Bias-Active-Learning}{https://github.com/drimpossible/Sampling-Bias-Active-Learning}}. 
\section{Results}\label{sec:results}

\begin{table}[t]
\small\addtolength{\tabcolsep}{-4.5pt}
\centering
\resizebox{1.05\columnwidth}{!}{
\begin{tabular}{|l|c|c|c|c|c|c|}
\hline
Dsets & Limit & FTZ ($\cap Q$) & MNB ($\cap Q$) & FTZ ($\cap S$) & MNB ($\cap S$) \\ \hline
SGN & 1.61  & $1.56 \pm 0.03$ & $1.15 \pm 0.32$ & $1.59 \pm 0.01$ & $1.57 \pm 0.01$ \\ \hline 
DBP & 2.64 & $2.50 \pm 0.02$ & $2.27 \pm 0.11$ & $2.51 \pm 0.0$ & $2.58 \pm 0.01$ \\ \hline
YHA & 2.30 &  $2.25 \pm 0.01$ &  $2.22 \pm 0.02$ & $2.25 \pm 0.0$ & $2.28 \pm 0.0$ \\ \hline 
YRP & 0.69 &  $0.69 \pm 0.0$ &  $0.56 \pm 0.13$ & $0.69 \pm 0.0$ & $0.69 \pm 0.01$\\ \hline 
YRF & 1.61 &  $1.56 \pm 0.02$ & $1.42 \pm 0.21$ & $1.56 \pm 0.0$ & $1.57 \pm 0.01$ \\ \hline 
AGN & 1.39 & $1.33 \pm 0.04$ & $1.13 \pm 0.17$ & $1.33 \pm 0.0$ & $1.35 \pm 0.01$ \\ \hline 
AMZP & 0.69 & $0.69 \pm 0.0$ &  $0.69 \pm 0.0$ & $0.69 \pm 0.0$ & $0.69 \pm 0.0$ \\ \hline 
AMZF & 1.61 &  $1.58 \pm 0.02$ &  $1.6 \pm 0.01$ & $1.59 \pm 0.0$ & $1.61 \pm 0.0$\\ \hline
\end{tabular}}
\caption{Label entropy with a large query size ($b=9$ queries).  $\cap Q$ denotes averaging across queries of a single run, $\cap S$ denotes the label entropy of the final collected samples, averaged across seeds. Naive Bayes ($\cap Q$) has biased (inefficient) queries  while FastText ($\cap Q$) shows stable, high label entropy showing a rich diversity in classes despite the large query size. Overall, the resultant sample  ($\cap S$) becomes balanced in both cases.}
\label{tab:label_entropy}
\end{table}


In this section, we study several aspects of sampling bias (class bias, feature bias) and the impact of relevant algorithmic factors (initial set selection, query size and query strategy. 

We evaluated the actively acquired queries and sample set for sampling bias, and for the stability as measured by \%intersection of collected sets across a critical influencing factor. Higher sample intersections indicate more stability increase to the chosen influencing factor. 

\subsection{Aspects of Sampling Bias}\label{sec:biases}

We study two types of sampling biases: (a) Class Bias and (b) Feature Bias.

\subsubsection{Class Bias}
Greedy uncertainty based query strategies are said to pick disproportionately from a subset of classes per query \cite{sener2018active, ebert2012ralf}, developing a lopsided representation in each query. However, its effect on the resulting sample set is not clear. We test this by measuring the Kullback-Leibler (KL) divergence between the ground-truth label distribution and the distribution obtained per query as one experiment ($\cap Q$), and over the resulting sample ($\cap S$) as the second. Let us denote $P$ as the true distribution of labels, $\hat{P}$ the sample distribution and $C$ the total number of classes. Since $P$ follows a uniform distribution, we can use Label entropy instead ($ L = -KL(P||\hat{P}) + log(C)$). Label entropy $L$ is an intuitive measure. The maximum label entropy is reached when sampling is uniform, $\hat{P}(x)=P(x)$, i.e. $L = log(C)$.


We present our results in Table \ref{tab:label_entropy}. We observe that across queries ($\cap Q$), FTZ with entropy strategy has a balanced representation from all classes (high mean) with a high probability (low std) while Multinomial Naive Bayes (MNB) results in more biased queries (lower mean) with high probability (high std) as studied previously.
However, we did not find evidence of class bias in the resulting sample ($\cap S$) in both models: FastText and Naive Bayes (column 5 and 6 from Table \ref{tab:label_entropy}). 

We conclude that entropy as a query strategy can be robust to class bias even with large query sizes. 

\subsubsection{Feature Bias}
Uncertainty sampling can lead to undesirable sampling bias in feature space \cite{settles2009active} by repeating redundant samples and picking outliers \cite{zhu2008active}. Diversity-based query strategies \cite{sener2018active} are used to address this issue, by selecting a representative subset of the data. In the context of active classification, it is good to pick the most informative samples to be the ones closer to class boundaries\footnote{In this work, we assume ergodicity in the setup. We do not consider incremental, online modeling scenarios where new modes or new classes are sequentially encountered.}. 

\begin{figure*}[!th]
\begin{subfigure}[h]{0.33\linewidth}
\includegraphics[width=\linewidth]{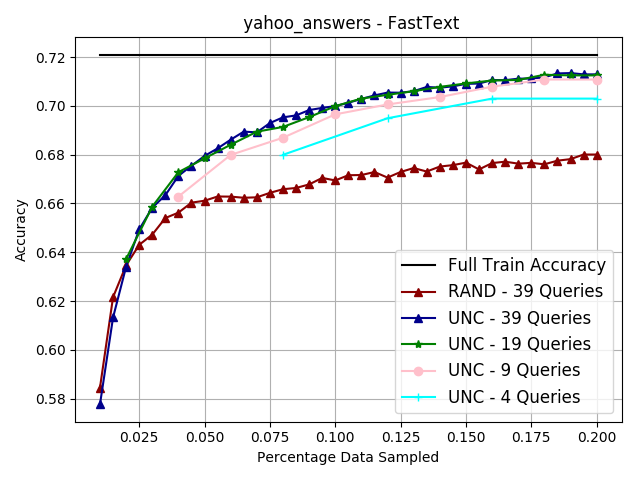}
\end{subfigure}
\begin{subfigure}[h]{0.33\linewidth}
\includegraphics[width=\linewidth]{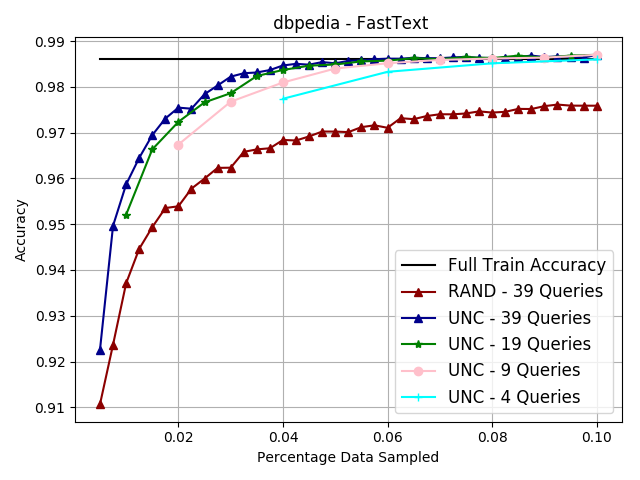}
\end{subfigure}
\begin{subfigure}[h]{0.33\linewidth}
\includegraphics[width=\linewidth]{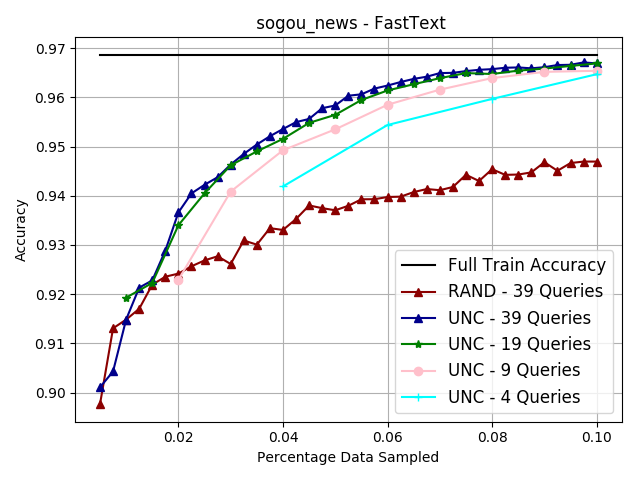}
\end{subfigure}
\begin{subfigure}[h]{0.33\linewidth}
\includegraphics[width=\linewidth]{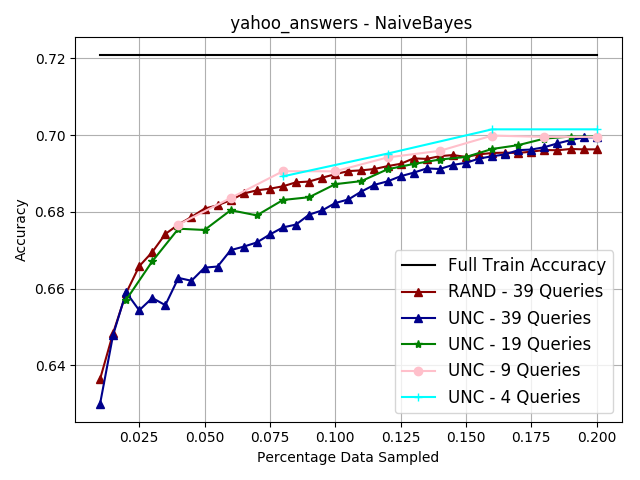}
\end{subfigure}
\begin{subfigure}[h]{0.33\linewidth}
\includegraphics[width=\linewidth]{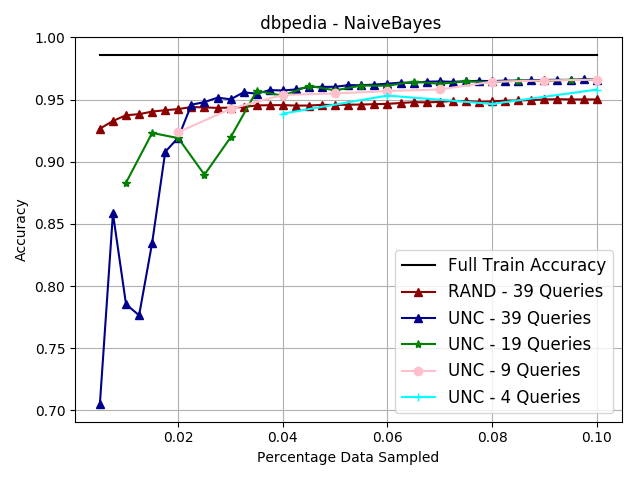}
\end{subfigure}
\begin{subfigure}[h]{0.33\linewidth}
\includegraphics[width=\linewidth]{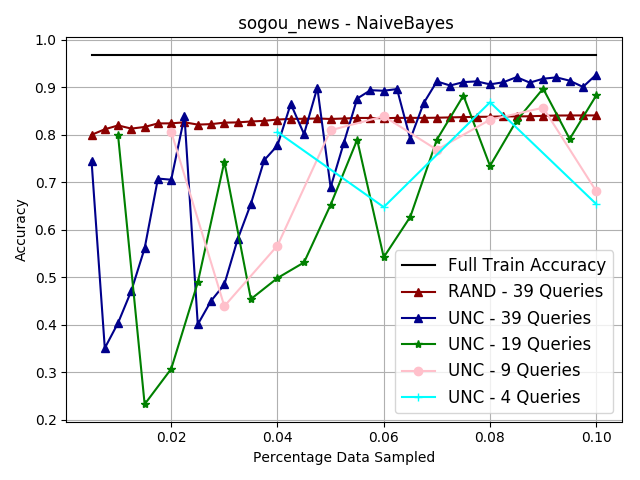}
\end{subfigure}
\caption{Accuracy across different number of queries $b$ for FastText and Naive Bayes, with  $b \times K$ constant. FastText is robust to increase in query size and significantly outperforms random in all cases. Naive Bayes: (Left) All including $b$=39 perform worse than random, (Center) All including b=$9$ eventually perform better than random (Right) $b=39$ performs better than random but larger query sizes perform worse than random. Uncertainty sampling with Naive Bayes suffers from sampling size bias.} 
\label{fig:sampling_biases}
\end{figure*}

Indeed, recent work suggests that the learning in deep classification networks may focus on small part of the data closer to class boundaries, thus resembling support vectors \cite{xu2018convergence, toneva19example}. To investigate whether uncertainty sampling also exhibits this behavior, we perform below a direct comparison with support vectors from a SVM. For this, we train a FTZ model on the full training data and train a SVM on the resulting features (sentence embeddings) to obtain the support vectors and compute the intersection of support vectors with each selected set. The percentage intersections are shown in Table \ref{tab:support}. The high percentage overlap is a surprising result which shows that the sampling is indeed biased but in a desirable way. Since the support vectors represent the class boundaries, a large percentage of selected data consists of samples around the class boundaries. This overlap indicates that the actively acquired training sample covers the support vectors well which are important for good classification performance. The overlap with the support vectors of an SVM (a fixed algorithm) also suggests that uncertainty sampling using deep models might generalize beyond FastText, to other learning algorithms. 
\begin{table}[t]
\small\addtolength{\tabcolsep}{-3pt}
\centering
  \begin{tabular}{|l|r|r|r|}
  \hline
  Dsets & Common\% & Chance\% & $\#SV$  \\ \hline
  SGN & $71.3 \pm 0.5$ & $9.3 \pm 0.5$ & 13184  \\ \hline 
  DBP & $86.3 \pm 0.5$ & $9.7 \pm 0.5$ & 1479   \\ \hline
  YRP & $57.3 \pm 0.5$ & $9.7 \pm 0.5$ & 31750    \\\hline
  AGN  & $45.0 \pm 0.8$ & $21.0 \pm 1.6$ & 1032  \\ \hline 
  \end{tabular}
\caption{Proportion of Support Vectors intersecting with our actively selected set calculated by $\frac{|\mathcal{S}_{SV}\cap \hat{\mathcal{S}_b}|}{|\mathcal{S}_{SV}|}$. Actively selected sets share large overlap with supports of an SVM (critical for classification).} 
\label{tab:support}
\vspace{-0.8cm}
\end{table}
\begin{table}[t]
\small\addtolength{\tabcolsep}{-4pt}
\centering
\begin{tabular}{|l|c||c|c||c|c|}
\hline
Dsets & Chance & FTZD & FTZS & MNBD & MNBS\\ \hline
SGN & 0.8 &  77.8 & 81.0 & 55.5 & 100.0 \\ \hline 
DBP & 0.9 &  79.7 & 81.3  & 79.7 & 100.0 \\ \hline
YHA & 3.7 & 69.0 & 73.6  & 89.5 & 100.0 \\ \hline 
YRP & 0.9 &  42.9 & 43.7  & 16.0 & 100.0 \\ \hline 
YRF & 3.6 &  67.7 & 71.6  & 13.6 & 100.0 \\ \hline 
AGN & 3.7 &  68.7 & 70.1  & 79.8 & 100.0 \\ \hline 
AMZP & 0.9 &  48.4 & 48.8  & 15.0 & 100.0 \\ \hline 
AMZF & 3.6 &  56.8 & 63.1  & 57.8 & 100.0 \\ \hline
\end{tabular}
\caption{\% Intersection of samples obtained with different seeds (ModelD) compared to same seeds (ModelS) and chance intersection for $b=39$ queries. We see that FastText is initialization independent (FTZD $\approx$ FTZS  $\gg$ Chance). NaiveBayes shows significant dependency on the initial set sometimes, while other times performs comparable to FastText.} 
\label{tab:seed_bias}
\end{table}
\begin{table*}[t]
\vspace*{-0.2cm}
\small\addtolength{\tabcolsep}{-4pt}
\centering
\begin{tabular}{|l|c||c|c||c|c|}
\hline
Dsets & Chance & FTZ $9 \cap 19 \cap 39$ & FTZ $39 \cap 39  \cap 39$ & MNB $9 \cap 19  \cap 39$ & MNB $39  \cap 39 \cap 39$\\ \hline
SGN & 0.83 $\pm 0.0$ & $77.0 \pm 0.5$ & $77.9 \pm 0.2$ & $31.9 \pm 0.0$ & $55.5 \pm 0.0$ \\ \hline 
DBP & 0.9 $\pm 0.0$& $80.0 \pm 0.1$ & $79.6 \pm 0.2$ & $82.3 \pm 0.0$  & $79.7 \pm 0.0$ \\ \hline
YHA & 3.7 $\pm 0.0$& $68.3 \pm 0.1$ & $69.0 \pm 0.0$ & $92.1 \pm 0.0$ & $89.5 \pm 0.0$ \\ \hline 
YRP & 0.9 $\pm 0.0$& $46.0 \pm 0.9$ & $42.7 \pm 1.0$ & $10.8 \pm 0.0$ & $16.0 \pm 0.0$\\ \hline 
YRF & 3.6 $\pm 0.0$& $68.4 \pm 0.2$ & $67.6 \pm 0.1$ & $14.2 \pm 0.0$ & $13.6 \pm 0.0$ \\ \hline 
AGN & 3.7 $\pm 0.0$& $70.3 \pm 0.2$ & $68.7 \pm 0.1$ & $81.6 \pm 0.0$ & $79.8 \pm 0.0$ \\ \hline 
AMZP & 0.9 $\pm 0.0$& $45.8 \pm 0.1$ & $48.2 \pm 0.2$ & $11.5 \pm 0.0$ & $15.0 \pm 0.0$\\ \hline 
AMZF & 3.6 $\pm 0.0$& $55.2 \pm 0.4$ & $57.0 \pm 0.2$ & $28.4 \pm 0.0$ & $57.8 \pm 0.0$\\ \hline
\end{tabular}
\caption{Intersection of samples obtained with different values of $b$. We see the intersection of samples selected with different number of intersections comparable to highest possible (different seeds) in FastText, far higher compared to chance intersection. This indicates similar samples are selected regardless of sample size. NaiveBayes does not show clear trends but occasionally the queried percentage drops significantly when increasing iterations, occasionally it remains unaffected.} 
\label{tab:sampling_bias}
\end{table*}

\textbf{Experimental Details:} We used a fast GPU implementation for training an SVM with a linear kernel \cite{wen2018thundersvm} with  default hyperparameters. Please refer to supplementary material for additional details. We ensured the SVM achieves similar accuracies as original FTZ model.

\subsection{Algorithmic Factors}\label{sec:causes}

We analyze three algorithmic factors of relevance to sampling bias: (a) Initial set selection (b) Query size, and, (c) Query strategy.
\begin{table*}[t]

\resizebox{\textwidth}{!}{
\begin{tabular}{|l|c||c|c||c|c|c|}
\hline
Dsets & Chance &  FTZ Ent-Ent  &  FTZ Ent-LC & FTZ Ent-DelEnt & FTZ DelEnt-DelLC & FTZ DelEnt-DelEnt\\ \hline
 SGN & $9.4 \pm 0.0$ & $84.6 \pm 0.2$  & $83.1 \pm 0.3$ & $81.7 \pm 0.1$ & $82.6 \pm 0.1$ & $84.2 \pm 0.1$\\ \hline 
  DBP & $9.3 \pm 0.0$ & $85.7 \pm 0.2$  & $85.5 \pm 0.3$  & $83.3 \pm 0.1$ & $83.0 \pm 0.4$ & $83.2 \pm 0.2$ \\ \hline 
  YHA & $19.0 \pm 0.0$& $79.0 \pm 0.0$ & $71.6 \pm 0.2$ & $76.3 \pm 0.1$ & $69.6 \pm 0.7$ & $75.6 \pm 3.9$  \\ \hline 
  YRP & $9.3 \pm 0.0$ & $58.4 \pm 0.6$ & $59.0 \pm 0.3$  & $59.0 \pm 0.6$ & $61.6 \pm 0.7$ & $62.1 \pm 0.1$ \\ \hline 
  YRF & $19.0 \pm 0.0$ & $77.8 \pm 0.2$  & $66.6 \pm 0.3$  & $75.8 \pm 0.1$ & $65.4 \pm 0.3$ & $80.1 \pm 0.2$\\ \hline 
  AGN  & $19.1 \pm 0.0$ & $78.3 \pm 0.1$  & $77.3 \pm 0.1$ & $77.1 \pm 0.3$ & $78.2 \pm 0.4$ & $79.0 \pm 0.3$ \\ \hline 
  AMZP & $9.5 \pm 0.0$ & $63.5 \pm 0.2$  & $63.5 \pm 0.3$  & $66.1 \pm 0.4$ & $70.0 \pm 0.1$ & $70.0 \pm 0.1$\\ \hline 
  AMZF & $19.0 \pm 0.0$ & $70.3 \pm 0.1$  & $64.3 \pm 0.2$ & $69.6 \pm 0.1$ & $65.6 \pm 0.2$ & $72.6 \pm 0.2$\\ \hline 
\end{tabular}}
\caption{Intersection of query strategies across acquisition functions. We observe that the \% intersection among samples in the Ent-LC is comparable to those Ent-Ent. Similarly, the Ent-DelEnt (entropy with deletion) is comparable to both DelEnt-DelLC and DelEnt-DelEnt showing robustness of FastText to query functions (beyond minor variation). DelEnt-DelEnt obtains similar intersections as compared to Ent-Ent, showing the robustness of the acquired samples to deletion.} 
\label{tab:query_acquisition}
\end{table*}

\subsubsection{Initial Set Selection}
To investigate the dependence of the actively acquired train set on the initial set, we compare the overlap (intersection) of the incrementally constructed sets from different random initial sets versus the same initial set. The results are shown in Table \ref{tab:seed_bias}. We first observe that chance overlaps (column 2) are very low - less than 4\%. Columns 3 and 5 present overlaps from different initial sets while 4 and 6 from same initial sets. We note from column 4 and 6 that due to the stochasticity of training in FTZ, we expect non-identical final sets even with same initial samples as well. The results demonstrate that samples obtained using FastText are largely initialization independent (low variation between columns 3 and 4) consistently across datasets while the samples obtained with Naive Bayes can be vastly different showing relatively heavy dependence on the initial seed. This indicates the relative stability of train set obtained with the posterior uncertainty of the actively trained FTZ as an acquisition function.

\subsubsection{Query size}
\label{subsubsec:querysize}
Since the sampled data is sequentially constructed by training models on previously sampled data, large query sizes  were expected to impact samples collected by uncertainty sampling and the performance thereof \cite{hoi2006large}. We experiment with various query sizes - (0.25\%, 0.5\%, 1\%) for DBP, SGN, YRP and AMZP and (0.5\%, 1\%, 2\%) for the rest corresponding to 9, 19 and 39 iterations. Figure \ref{fig:sampling_biases} shows that FastText (top row) has very stable performance across sample sizes while MNB (bottom row) show more erratic performance. Table \ref{tab:sampling_bias} presents the intersection of samples obtained with different query sizes across multiple runs. We observe a high overlap of the acquired samples across different query sizes indicating that the performance is independent of the query size (compare column 3 to column 4 where the size is held constant) while MNB results in lower overlap with more erratic behavior due to change in the query size (compare column 5 compared to column 6). 

\subsubsection{Query strategy}
\label{subsubsec:qstrategy}
We now investigate the impact of various query strategies using FastText by evaluating and comparing the correlation between the respective actively selected sample sets.

\textbf{Acquisition Functions:} We compare four uncertainty query strategies: Least Confidence (LC) and Entropy (Ent), with and without deletion of least uncertain samples from the training set. Deletion of least uncertain samples reduces the dependence on the initial randomly selected set. The results are presented in Table \ref{tab:query_acquisition}. We present five of the ten possible combinations and again observe the high degree of overlap in the collected samples. It can be concluded that the approach is fairly robust to these variations in the query strategy.

\textbf{Ensembles versus Single Models:} A similar experiment was conducted to investigate the overlap between a single FTZ model and a probabilistic committee of models (5-model ensemble with FTZ \cite{lakshminarayanan2017simple}) to identify comparative advantages of  using ensemble methods. The results are presented in Table \ref{tab:query_model} showing little to no difference in sample overlaps. \footnote{The ensembles were too costly to run on larger datasets, so the results for YHA, AMZP and AMZF could not be obtained.} We conclude that more expensive sampling strategies commonly used, like ensembling, may offer little benefit compared to using a single FTZ model with posterior uncertainty as a query function.

 The experiments in this section demonstrate that uncertainty based sampling using deep models like FTZ show no class bias or an undesirable feature bias (and favorable bias to class boundaries). There is also a high degree of robustness to algorithmic factors, especially query size, a surprisingly high degree of overlap in the resulting training samples and stable performances (classification accuracy). Additionally, all uncertainty query strategies perform well, and expensive sampling strategies like ensembling offer little benefit. We conclude that sampling biases demonstrated in active learning literature do hold well with traditional models, however, they do not seem to translate to deep models like FTZ using (posterior) uncertainty.

\begin{table}[t]
\resizebox{1.05\columnwidth}{!}{
\begin{tabular}{|l|c|c|c|c|c|}
\hline
Dsets & Chance & \makecell{FTZ-FTZ\\Ent} & \makecell{FTZ-5F\\TZ Ent} & \makecell{5FTZ-5FTZ\\Ent-LC} & \makecell{5FTZ-5FTZ\\Ent-Ent}  \\ \hline
SGN & $9.4 \pm 0.0$ & $84.6 \pm 0.2$ & $86.3 \pm 0.2$ & $85.4 \pm 0.4$ & $85.8 \pm 0.0$ \\ \hline 
DBP & $9.3 \pm 0.0$ & $85.7 \pm 0.2$  & $86.6 \pm 0.3$ & $86.78 \pm 0.1$ & $87.8 \pm 0.2$ \\ \hline
YRP & $9.3 \pm 0.0$ & $58.4 \pm 0.6$ & $58.1 \pm 0.7$ & $58.3 \pm 0.3$ & $58.2 \pm 0.2$\\ \hline 
YRF & $19.0 \pm 0.0$ & $77.8 \pm 0.2$ & $79.0 \pm 0.3$ & $68.5 \pm 1.1$ & $77.6 \pm 0.3$  \\ \hline 
AGN & $19.1 \pm 0.0$ & $78.3 \pm 0.1$ & $79.0 \pm 0.2$ & $79.1 \pm 0.2$ & $77.9 \pm 0.2$ \\ \hline 
\end{tabular}}
\caption{Intersection of query strategies across single and ensemble of 5FTZ models. We observe that the \% intersection of samples selected by ensembles and single models is comparable to intersection among either. The 5-model committee does not seem to add any additional value over selection by a single model.} 
\label{tab:query_model}
\end{table}

 \begin{table*}[t]
\small\addtolength{\tabcolsep}{-3pt}
\centering
  \begin{tabular}{|l||r|r|r||r|r|}\hline
  Dsets & Chance & FTZ-Ent-Ent & FTZ Ent-LC & SV Chce\% & SV Com\% \\ \hline
  TQA & $15.1 \pm 0.0$ &  $\mathbf{59.7 \pm 0.5}$  & $\mathbf{56.3 \pm 1.4}$ & $18.7 \pm 6.1$  & $\mathbf{79.0 \pm 3.6}$  \\ \hline
  \end{tabular}
  \caption{Results of sample selection from previous investigations on small datasets (Trec-QA).} 
  \label{tab:trecqa_stats}
\end{table*}
\section{Application: Active Text Classification}
\label{sec:comparison}
Experimental results from the previous sections suggest that entropy function with a single FTZ model would be a good baseline for active text classification. We compare our baseline with the latest work in deep active learning for text classification - BALD \cite{siddhant2018deep} and with the recent diversity based Coreset query function \cite{sener2018active} which uses a costly K-center algorithm to build the query. Experiments are performed on TREC-QA for a fair comparison (used by \cite{siddhant2018deep}).  Table \ref{tab:trecqa_stats} shows that the results of our study generalize to small datasets like TREC-QA.

\begin{figure}[t]
\centering
\includegraphics[width=1.0\linewidth]{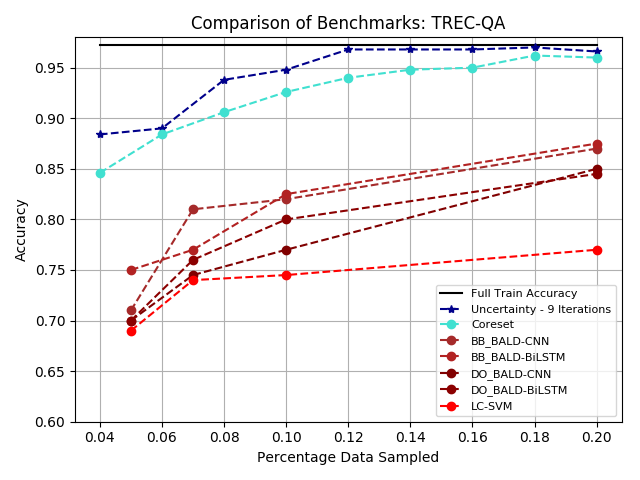}
\caption{Active text classification: Comparison with K-Center Coreset, BALD and SVM algorithms. Accuracy is plotted against percentage data sampled. We reach full-train accuracy using 12\% of the data, compared to BALD which requires 50\% data and perform significantly worse in terms of accuracy. We also outperform K-center greedy Coreset at all sampling percentages without utilizing additional diversity-based augmentation.} 
\label{fig:comparison}
\vspace{-0.4cm}
\end{figure}

\begin{table*}[t]
\centering
\resizebox{\textwidth}{!}{
\begin{tabular}{|l|c|c|c|c|c|c|c|c|}\hline
Model & AGN & DBP & SGN & YRF & YRP & YHA & AMZP & AMZF \\ \hline  
VDCNN \cite{conneau2016very} & 91.3 & 98.7 & 96.8 & 64.7 & 95.7 & 73.4 & 95.7 & 63.0 \\ \hline
DPCNN \cite{johnson2017deep} & 93.1 & \textbf{99.1} & \textbf{98.1} & \textbf{69.4 } & \textbf{97.3} & \textbf{76.1} & \textbf{96.7} & \textbf{65.2} \\ \hline
WC-Reg \cite{qiao2018anew} & 92.8 & 98.9 & \textbf{97.6} & 64.9 & 96.4 & 73.7 & 95.1 & 60.9 \\ \hline
DC+MFA \cite{wang2018densely} & 93.6 & \textbf{99.2} & -  & 66.0 & 96.5 & - & - & 63.0 \\ \hline
DRNN \cite{wang2018disconnected} & \textbf{94.5} &\textbf{ 99.2} & - & \textbf{69.1}  & \textbf{97.3} & 70.3 &  \textbf{96.5} & \textbf{64.4}  \\ \hline
ULMFiT \cite{howard2018universal} & \textbf{95.0} & \textbf{99.2} & - & \textbf{70.0} & \textbf{97.8} & - & -  & - \\ \hline
EXAM \cite{du2019explicit} & 93.0 & 99.0 & - & - & - & 74.8 &  95.5 & 61.9  \\ \hline
Ours: ULMFiT (Small data) & 93.7 (20) & \textbf{99.2 (10)} & 97.0 (10) & 67.6 (20) & \textbf{97.1 (10)} & 74.3 (20) & \textbf{96.1 (10)} & \textbf{64.1 (20)} \\ \hline
Ours: ULMFiT (Tiny data) & 91.7 (8) & 98.6 (2.3) & \textbf{97.4 (6.3)} & 66.3 (8) & 96.7 (4) & 73.3 (8) & 95.8 (4) & 62.9 (8) \\ \hline
\end{tabular}}
\caption{Comparison of accuracies with state-of-the-art approaches (earliest-latest) for text classification (\%dataset in brackets). We are competitive with state-of-the-art models while using 5x-40x compressed datasets. }
\label{tab:comprehensive}
\end{table*}
\begin{table}[t]
\centering
\resizebox{\columnwidth}{!}{
\begin{tabular}{|l|c|c|c|c|}\hline
ULMFiT & AGN & DBP & YRP & YRF \\ \hline  
Full & \textbf{95.0} & \textbf{99.2} & \textbf{97.8} & \textbf{70.0} \\ \hline
Ours-Small & 93.7 (20) & \textbf{99.2 (10)} & \textbf{97.1 (10)} & 67.6 (20) \\ \hline
Ours-Tiny & 91.7 (8) &  \textbf{98.6 (2.3)} & 96.7 (4) & 66.3(8) \\ \hline
\end{tabular}}
\caption{ULMFiT: Resulting sample $\hat{\mathcal{S}}_b$ compared to reported accuracies in \cite{howard2018universal} (\%dataset in brackets). We observe that using our cheaply obtained compressed datasets, we can achieve similar accuracies with 25x-200x speedup (5x less epochs, 5x-40x less data). Transferability to other models is evidence of the generalizability of the subset collected using FTZ to other deep models.}
\label{tab:ulmfitonly}
\end{table}
The results are shown in Figure \ref{fig:comparison} using the baseline with the query size of 2\% of the full dataset (b=9 queries). Note that uncertainty sampling converges to full accuracy using just 12\% of the data, whereas \cite{siddhant2018deep} required 50\% of the data. There is also a remarkable accuracy improvement over \cite{siddhant2018deep} which can be largely attributed to the models used (FastText versus 1-layer CNN/BiLSTM). Also, uncertainty sampling outperforms diversity-based augmentations like Coreset Sampling \cite{sener2018active} before convergence. Thus, we establish a new state-of-the-art baseline for further research in deep active text classification.

\section{Application: Training of Large Models}
\label{sec:implications}

The cost and time needed to get and label vast amounts of data to train large DNNs is a serious impediment to creating new and/or better models. Our study suggests that the training samples collected with uncertainty sampling (entropy) on a single model FTZ may provide a good representation (surrogate) for the entire dataset. Buoyed by this, we investigate if we can speedup training of ULMFiT \cite{howard2018universal} using the surrogate dataset. We show these results in Table \ref{tab:ulmfitonly}. We achieve 25x-200x speedup\footnote{The cost of acquiring the training data using FTZ-Ent is negligible in comparison.} (5x fewer epochs, 5x-40x smaller training size). We also benchmark the performance against the state-of-the-art on text classification as shown in Table \ref{tab:comprehensive}. We conclude that we can significantly compress the training datasets and speedup classifier training time with little tradeoff in accuracy.

\textbf{Implementation Details:} We use the official github repository for ULMFiT\footnote{\url{https://github.com/fastai/fastai/tree/master/courses/dl2/imdb\_scripts}}, use default hyperparameters and train on one NVIDIA Tesla V100 16GB GPU. Further details are provided in supplementary material. 
\section{Related Work}\label{sec:related}
We now expand on the brief literature review in Section \ref{sec:introduction} to better contextualize our work. We divide the past works into (i) Traditional Models and (ii) Deep Models.

\textbf{Sampling Bias in Classical AL in NLP:} Active learning (AL) in text classification started with greedy uncertainty query strategy from a pool using decision trees \cite{lewis1994sequential}, which was shown to be effective and led to widespread adoption with classifiers like SVMs \cite{tong2001support}, Naive Bayes \cite{roy2001toward} and KNN \cite{fujii1998selective}. This strategy was also applied to other NLP tasks like parse selection \cite{baldridge2004active}, sequence labeling \cite{settles2008analysis} and information extraction \cite{thompson1999active}. These early papers popularized two greedy uncertainty query methods: Least Confident and Entropy. 

Issues of lack of diversity (large reduduncy in sampling) \cite{zhang2000value} and lack of robustness (high variance in sample quality)\cite{krogh1994nnensembles} guided subsequent efforts. The two most popular directions were: (i) augmenting uncertainty with diversity measures \cite{hoi2006large, brinker2003incorporating, tang2002active} and (ii) using query-by-committee \cite{mccallum1998employing, liere1997active}. For a comprehensive survey of classical AL methods for NLP, please refer to \cite{settles2009active}.

\textbf{Sampling Bias in Deep AL:} Deep active learning approach adapt the above framework to the training of DNNs on large data. Two main query strategies are used: (i) ensemble based greedy uncertainty, which represents a probabilistic query-by-committee paradigm \cite{gal2017deep, beluch2018power}, and (ii) diversity based measures \cite{sener2018active,ducoffe2018adversarial}. Papers proposing diversity based approaches find that greedy uncertainty based sampling (using ensemble and single model) perform significantly worse than random (See Figures 4 and 2 respectively in \cite{sener2018active,ducoffe2018adversarial}). They attribute the poor performance to redundant, highly correlated sampling selected using uncertainty based methods and justify the need for  prohibitively expensive diversity-based approaches (Refer section 2 of \cite{sener2018active} for details on the expensiveness of various diversity sampling methods). However, K-center greedy coreset sampling scales poorly: we were only able to use it on TREC-QA (a small dataset). On the other hand, ensemble-based greedy uncertainty methods find that probabilistic averaging from a committee \cite{gal2017deep,beluch2018power} performs better than single model as with on diversity based methods like coreset\cite{gissin2019discriminative, beluch2018power}. Current approaches in text classification literature mostly adopt the ensemble based greedy uncertainty framework \cite{siddhant2018deep, lowell2018transferable, zhang2017active}. 

However, our work demonstrates the problems of sampling bias and efficiency may not translate from shallow to deep approaches.  Recent evidence from image domain \cite{gissin2019discriminative} demonstrates atleast a subset of our findings generalize to other DNNs (class bias and query functions). Uncertainty sampling using a deep model like FTZ demonstrates surprisingly good sampling properties without using ensembles or bayesian methods. Ensembles do not seem to significantly affect sampling. Whether this behavior generalizes to other deep models and tasks is yet to be seen. 

\textbf{Other Related Works:} An interesting set of papers \cite{soudry2018implicit, xu2018convergence} show that deep neural networks trained with SGD converge to the maximum margin solution in the linearly separable case. Several works investigate the possibility that deep networks give high importance to a subset of the training dataset \cite{toneva19example, vodrahalli2018all, birodkar2019semantic}, resembling supports in support vector machines. In our experiments, we find that active learning with uncertainty sampling with deep models like FTZ has a (surprisingly) large overlap with the support vectors of an SVM. Thus, it seems to have a inductive bias for class boundaries, similar to the above works. Whether this property generalizes to other deep models is yet to be seen. 
\section{Conclusion}\label{sec:conclusion}

We conducted a large empirical study of sampling bias and efficiency, along with algorithmic factors which impacting active text classification. We conclude that uncertainty sampling with deep models like FastText.zip exhibits negligible class bias, seems to be favorably biased to sampling data points near class boundaries, is robust to various algorithmic factors and expensive sampling strategies like ensembling offer little benefit. Also, we find a surprisingly large overlap of actively acquired points with supports of a SVM. We additionally show that uncertainty sampling can be effectively used to bootstrap the training of large DNN models by generating compact surrogate datasets (5x-40x compression). Finally, FTZ-Ent provides a strong baseline for deep active text classification, outperforming previous results by a margin of 4x less data.   

The current work opens up several directions for future investigations. To list a few: (a) a deeper look into the nature of sampled data - their distribution in the feature space, as well as their importance for the task at hand; (b) the creation of surrogate datasets for a variety of applications, including hyperparameter and architecture search, etc; (c) an extension to other deep models (beyond FTZ) and beyond classification models; and, (d) an extension to semi-supervised, online and continual learning.\\ 

\textbf{Acknowledgements}: We thank Prof. Vineeth Balasubramium, IIT Hyderabad, India for the many helpful suggestions and discussions. 

\newpage
\bibliography{emnlp2019}
\bibliographystyle{acl_natbib}
\clearpage
\appendix
In this document, we present statistics, additional tables and hyperparameters left out of the main work due to lack of space.

\section{Dataset Details}
Details of the train, test sizes and number of classes for each dataset can be found in Table \ref{tab:dataset_details}.
\begin{table}[t]
\centering
\small\addtolength{\tabcolsep}{-3pt}
\begin{tabular}{|l|c|c|c|c|c|c|c|c|}\hline
 & AGN & SGN & DBP & YHA & YRP & YRF & AMZP & AZMF \\ \hline
\#Class & 4 & 5 & 14 & 10 & 2 & 5 & 2 & 5 \\ \hline
\#Train& 120k & 450k & 560k & 1.4M & 560k & 650k & 3.6M & 3.0M\\ \hline
\#Test & 7.6k & 60k & 70k & 60k & 38k & 50k & 400k & 650k\\ \hline
\end{tabular}
\caption{Details about the dataset sizes (both train and test) along with the number of classes.}
\label{tab:dataset_details}
\end{table}

\section{Experiment Hyperparameters}
In this section, we detail the complete list of hyperparameters, for reproducibility. We will release our code on Github.

\subsection{Models}
We describe the model hyperparameters used for 4 models: (i) FastText (ii) SVM (iii) ULMFiT (iv) Multinomial Naive Bayes for reproducibility.
\subsubsection{FastText}
We use the original implementation\footnote{\url{https://github.com/facebookresearch/fastText/}}. The hyper-parameters used for each dataset can be found in Table \ref{tab:fast_hyper}. We chose to use the zipped version of FastText for optimized memory usage without loss of accuracy, or speed.

\begin{table}[t]
\begin{center}
\small\addtolength{\tabcolsep}{-3pt}
\begin{tabular}{|c|c|c|c|c|c|}
\hline
Dsets & Emb Dim & NGrams & Epochs & LR & Acc Full  \\ \hline
SGN &  25 & 2 & 10 & 0.25 & 96.9   \\ \hline
TQA &  25 & 2 & 20 & 0.75 & 97.2 \\ \hline
DBP  &  25 & 2 & 10 & 1 & 98.6 \\ \hline
YHA  &  25 & 2 & 10 & 0.02 & 72.1   \\ \hline
YRP  &  25 & 2 & 10 & 0.05 & 95.6 \\ \hline
YRF & 25 & 2 & 10 & 0.05 & 63.6 \\ \hline
AGN & 25 & 2 & 10 & 0.25 & 92.1 \\ \hline
AMZP &   25 & 2 & 10 & 0.01 & 94.2  \\ \hline
AMZF  & 25 & 2 & 10 & 0.01 & 59.6  \\ \hline
\end{tabular}
\caption{Hyperparameters Used for FastText: Embedding dimension, Number of n-grams, number of epochs, learning rate, accuracy obtained using the full train set}
\label{tab:fast_hyper}
\end{center}
\end{table}

\begin{table}[t]
\begin{center}
\small\addtolength{\tabcolsep}{-2pt}
\begin{tabular}{|c|c|c|c|c|c|c|c|c|c|}
\hline
Dsets & NLL & BrierL & ECE &  VarR & ENT & STD  \\ \hline
SGN & 0.14 &  0.01 & 0.01  & 0.02 & 0.07 & 0.39   \\ \hline
DBP  & 0.07 &  0.0 & 0.01 &  0.0 & 0.02 & 0.26 \\ \hline
YHA  & 1.37 &  0.05 & 0.16 &  0.12 & 0.5 & 0.27   \\ \hline
YRP  & 0.16 &  0.04 & 0.02  & 0.03 & 0.11 & 0.47 \\ \hline
YRF & 1.15 & 0.11 & 0.17 & 0.21 & 0.73 & 0.31 \\ \hline
AGN & 0.46 &  0.03 & 0.04 & 0.02 & 0.08 & 0.42 \\ \hline
AMZP & 0.26 &  0.05 & 0.04 & 0.02 & 0.08 & 0.48  \\ \hline
AMZF  & 1.32 & 0.12 & 0.21 & 0.22 & 0.77 & 0.31  \\ \hline
\end{tabular}
\caption{\label{font-table} Metrics measured after training FastText (FTZ-Ent) model on the resulting sample, with 39 queries, using entropy query strategy. We observe that NLL and Multiclass Brier Score remains low. The model is also well calibrated, i.e. gives calibrated uncertainty estimates.}
\label{tab:measures}
\end{center}
\end{table}
\begin{table*}[t]
\small\addtolength{\tabcolsep}{-4.5pt}
\begin{tabular}{|l|c||c|c|c|c||c|c|c|c||}
\hline
Dsets & Chance & FTZ Ent-Ent & FTZ Ent-LC & MNB Ent-Ent & MNB Ent-LC   & FTZ Ent-Ent & FTZ Ent-LC & MNB Ent-Ent & MNB Ent-LC \\ \hline
 SGN & $9.4 \pm 0.0$ & $81.6 \pm 0.1$  & $80.1 \pm 0.3$ & $52.9 \pm 0.0$ & $39.8 \pm 0.0$  & $74.8 \pm 0.3$  & $73.4 \pm 0.6$ & $35.0 \pm 0.0$ & $34.1 \pm 0.0$ \\ \hline 
  DBP & $9.3 \pm 0.0$ & $82.6 \pm 0.2$  & $82.2 \pm 0.1$ & $84.9 \pm 0.0$ & $69.8 \pm 0.0$ & $77.4 \pm 0.1$  & $76.6 \pm 0.2$ & $79.5 \pm 0.0$ & $64.1 \pm 0.0$ \\ \hline 
  YHA & $19.0 \pm 0.0$ & $75.0 \pm 0.1$  & $71.6 \pm 0.1$ & $90.9 \pm 0.0$ & $76.8 \pm 0.0$  & $66.1 \pm 0.1$ & $66.7 \pm 0.1$ & $86.4 \pm 0.0$ & $72.0 \pm 0.0$ \\ \hline 
  YRP & $9.3 \pm 0.0$ & $59.4 \pm 0.3$  & $59.6 \pm 0.4$ & $32.5 \pm 0.0$ & $32.5 \pm 0.0$  & $56.4 \pm 0.7$  & $56.4 \pm 0.6$ & $19.9 \pm 0.0$ & $19.9 \pm 0.0$\\ \hline 
  YRF & $19.0 \pm 0.0$ & $75.1 \pm 0.1$  & $62.0 \pm 0.1$ & $69.6 \pm 0.0$ & $60.7 \pm 0.0$  & $67.2 \pm 0.3$  & $53.6 \pm 0.1$ & $55.5 \pm 0.0$ & $44.8 \pm 0.0$\\ \hline 
  AGN  & $19.1 \pm 0.0$ & $75.8 \pm 0.3$  & $75.1 \pm 0.1$ & ?$81.1 \pm 0.0$ & $71.5 \pm 0.0$   & $70.6 \pm 0.2$  & $69.1 \pm 0.0$ & $76.2 \pm 0.0$ & $67.3 \pm 0.0$ \\ \hline 
  AMZP & $9.5 \pm 0.0$ & $60.2 \pm 0.1$  & $60.2 \pm 0.3$ & $32.2 \pm 0.0$ & $32.2 \pm 0.0$  & $52.7 \pm 0.6$ & $52.7 \pm 0.1$ & $23.5 \pm 0.0$ & $23.5 \pm 0.0$ \\ \hline 
  AMZF & $19.0 \pm 0.0$ & $64.8 \pm 0.3$ & $58.5 \pm 0.3$ & $64.2 \pm 0.0$ & $57.4 \pm 0.0$  & $55.2 \pm 0.1$  & $48.4 \pm 0.1$ & $55.2 \pm 0.0$ & $50.4 \pm 0.0$\\ \hline 
\end{tabular}
\caption{Intersection across query strategies using 19 and 9 iterations (mean $\pm$ std across runs) and different seeds} 
\label{tab:query_acquisition}
\end{table*}
\begin{table*}[t]
\small\addtolength{\tabcolsep}{-4.5pt}
  \centering
  \begin{tabular}{|l||r||r|r||r|r||r|r||}
  \hline 
  Datasets & Limit & FTZ ($\cap Q$) & MNB ($\cap Q$)  &  FTZ ($\cap Q$) &  MNB ($\cap Q$) & FTZ ($\cap Q$) & MNB ($\cap Q$) \\ \hline \hline
  SGN & 1.6 & $1.5 \pm 0.1$ & $1.4 \pm 0.2$ & $1.6 \pm 0.0$ & $1.3 \pm 0.2$  & $1.6 \pm 0.0$ & $1.2 \pm 0.2$  \\ \hline 
  DBP & 2.6 & $2.5 \pm 0.1$ & $2.3 \pm 0.1$ & $2.5 \pm 0.1$ & $2.3 \pm 0.2$  & $2.5 \pm 0.1$ & $2.3 \pm 0.1$ \\ \hline
  YA &  2.3 & $2.3 \pm 0.0$ &  $2.3 \pm 0.0$ &  $2.3 \pm 0.0$  &  $2.2 \pm 0.0$  &  $2.3 \pm 0.0$ &   $2.2 \pm 0.0$ \\ \hline 
  YRP &  0.7 &  $0.7 \pm 0.0$ &  $0.7 \pm 0.1$&  $0.7 \pm 0.0$ &  $0.6 \pm 0.2$ &  $0.7 \pm 0.0$ &  $0.7 \pm 0.0$ \\ \hline 
  YRF &  1.6 &  $1.6 \pm 0.0$ & $1.5 \pm 0.1$  & $1.6 \pm 0.0$ &  $1.4 \pm 0.2$ &  $1.6 \pm 0.0$ & $1.3 \pm 0.2$ \\ \hline 
  AGN  & 1.4 & $1.3 \pm 0.0$ & $1.3 \pm 0.1$ & $1.3 \pm 0.1$ & $1.1 \pm 0.2$ & $1.3 \pm 0.0$ & $1.1 \pm 0.1$ \\ \hline 
  AMZP  & 0.7 & $0.7 \pm 0.0$ &  $0.7 \pm 0.1$ & $0.7 \pm 0.0$ &  $0.7 \pm 0.0$ & $0.7 \pm 0.0$ &  $0.7 \pm 0.0$\\ \hline 
  AMZF &  1.6  &  $1.6 \pm 0.0$ &  $1.6 \pm 0.0$ &  $1.6 \pm 0.0$ &  $1.6 \pm 0.1$  &  $1.6 \pm 0.0$ &  $1.6 \pm 0.0$\\ \hline
  \end{tabular}
  \centering
\caption{Class Bias Experiments: Average Label entropy (mean $\pm$ std ) across  query iterations, for 39, 19 and 4 query iterations each.} 
 \label{tab:label_entropy}
\end{table*}
\subsubsection{ULMFiT}
For ULMFiT, we used the default hyperparameters from the author's implementation\footnote{\url{https://github.com/fastai/fastai/tree/master/courses/dl2/imdb\_scripts}}, except the batch size which we set to 32. We recall that ULMFiT has two steps: the fine-tuning of the language model and the fine-tuning of the classifier. We initialized the language model with the  pre-trained weights released by the authors. Results of a pre-training on Wikitext-103 consisting of 28,595 pre-processed Wikipedia articles and 103 million words. For each compressed datasets (small and very small), we fine-tuned the language model and the classifier for 10 epochs. For fine-tuning both language model and classifier, we used a NVIDIA Tesla V100 16GB.   

The hyperparameters for the language model are: batch size of $32$, learning rate of 4e-3, bptt of $70$, embedding size of $400$, $1150$ hidden units per hidden layer and $3$ hidden layers. Adam Optimizer with $\beta_1=0.8$ and $\beta_2=0.99$. The dropout rates are: 0.15 between LSTM layers, 0.25 for the input layer, 0.02 for the embedding layer, 0.2 for the internal LSTM recurrent weights.  
        
The hyperparameters for the classifier are: batch size of $32$, learning rate of $0.01$, embedding size of $400$, $1150$ hidden units per hidden layer and $3$ hidden layers. Adam Optimizer with $\beta_1=0.8$ and $\beta_2=0.99$. The dropout rates are: 0.3 between LSTM layers, 0.4 for the input layer, 0.05 for the embedding layer, 0.5 for the internal LSTM recurrent weights.  

\subsubsection{Multinomial Naive Bayes (MNB)}
We use the scikit-learn implementation of Multinomial Naive Bayes \footnote{\url{https://scikit-learn.org/stable/modules/generated/sklearn.naive_bayes.MultinomialNB.html}} with default hyperparameters: smoothing parameter $\alpha=1.0$, fit prior set to True and class prior set to None. As input to our MNB, we use the scikit-learn implementation of the TFIDF Vectorizer \footnote{\url{https://scikit-learn.org/stable/modules/generated/sklearn.feature_extraction.text.TfidfVectorizer.html}}. All default hyperparameters remain unchanged except that we use a maximum feature threshold of $50000$, we remove all stop words contained in the default list 'english' and we set sublinear tf to True.

\subsubsection{SVM}
To compute the support vectors of the datasets we used ThuderSVM, a Fast SVM library running on a V100 GPU. \footnote{\url{https://github.com/Xtra-Computing/thundersvm}}. We used the SVC with a linear kernel, degree = 3, gamma = auto, coef0 = 0.0, C = 1.0, tol = 0.001, probability = False, classweight = None, shrinking = False, cachesize = None, verbose = False, max iter = -1, gpuid=0, maximum memory size = -1, random state = None and decision function = 'ovo'. 


\section{Experiments}

\subsection{Class Bias}
We provide in Table \ref{tab:label_entropy} the complete results of our class bias experiments with 39, 19 and 4 iterations using entropy query strategy. 
\subsubsection{Results Across Iterations}
We provide in Table \ref{tab:query_acquisition} the results of our intersection experiments for 19 and 9 iterations using entropy query strategy for FastText (FTZ) and Multinomial Naive Bayes (MNB). 
\subsection{Metrics Affecting Uncertainty}
We provide in Table \ref{tab:measures} several metrics measured on the resulting samples of each dataset after 39 queries and using the entropy query strategy. NLL denotes the negative log-likelihood, BrierL denotes the Brier Score Loss, ECE denotes the expected calibration error, VarR denotes the variation ratio, ENT denotes the entropy, STD denotes the standard deviation. We measure these properties of the predicted sample and compute their average over the dataset. We observe that the FastText model is well calibrated except for YRF and AMZF. Similar trends are observed in the average uncertainty measures.
\subsection{Accuracy Plots for Remaining Datasets}
\begin{figure*}[h!]
\begin{subfigure}[h]{0.33\linewidth}
\includegraphics[width=\linewidth]{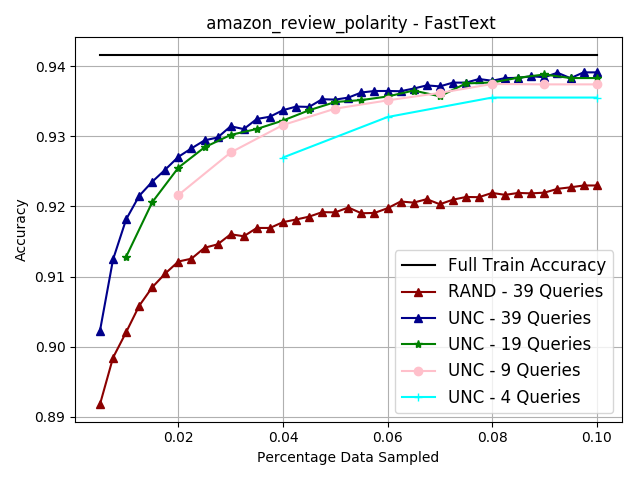}
\end{subfigure}
\begin{subfigure}[h]{0.33\linewidth}
\includegraphics[width=\linewidth]{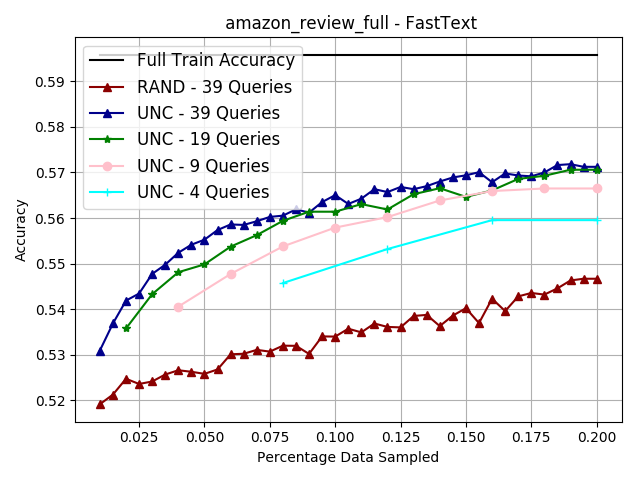}
\end{subfigure}
\begin{subfigure}[h]{0.33\linewidth}
\includegraphics[width=\linewidth]{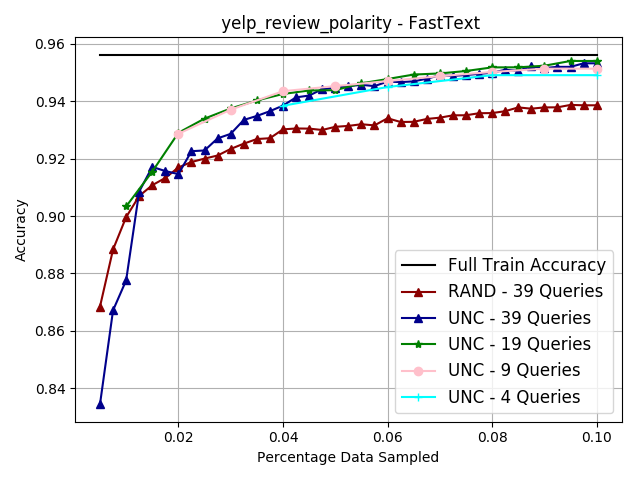}
\end{subfigure}
\begin{subfigure}[h]{0.33\linewidth}
\includegraphics[width=\linewidth]{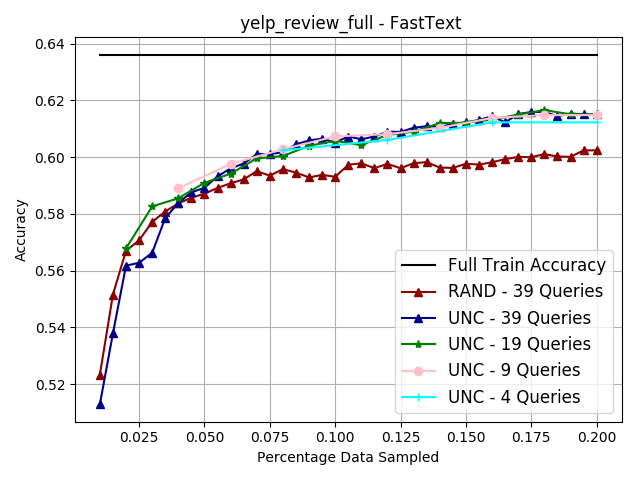}
\end{subfigure}
\begin{subfigure}[h]{0.33\linewidth}
\includegraphics[width=\linewidth]{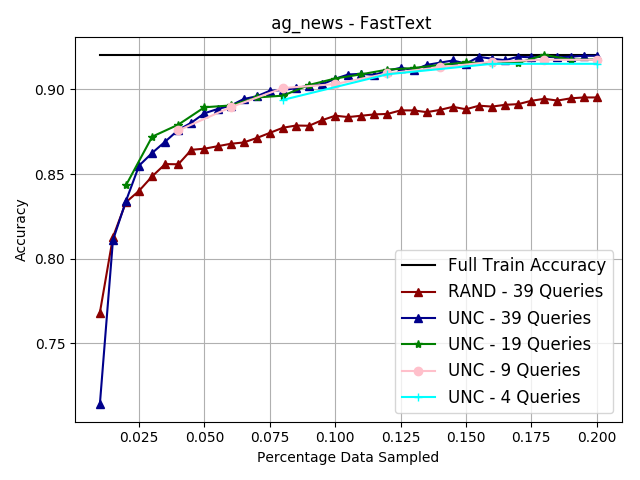}
\end{subfigure}
\begin{subfigure}[h]{0.33\linewidth}
\includegraphics[width=\linewidth]{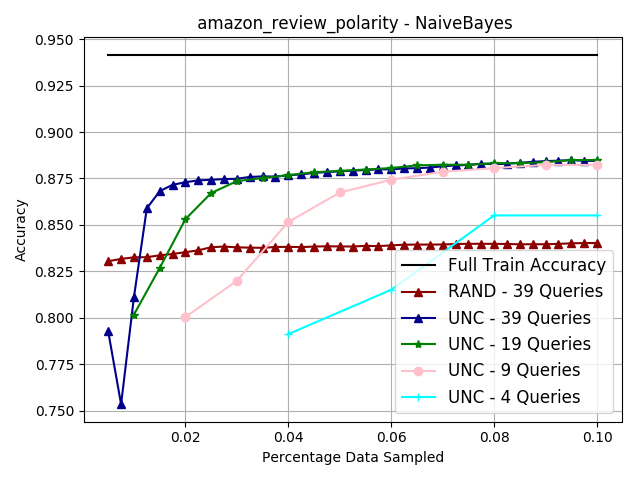}
\end{subfigure}
\begin{subfigure}[h]{0.33\linewidth}
\includegraphics[width=\linewidth]{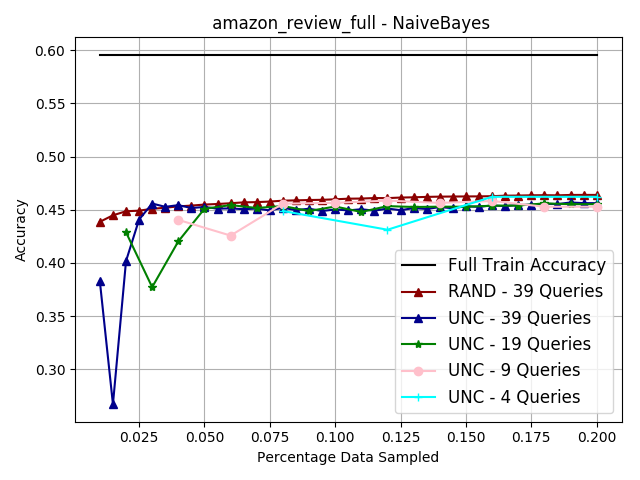}
\end{subfigure}
\begin{subfigure}[h]{0.33\linewidth}
\includegraphics[width=\linewidth]{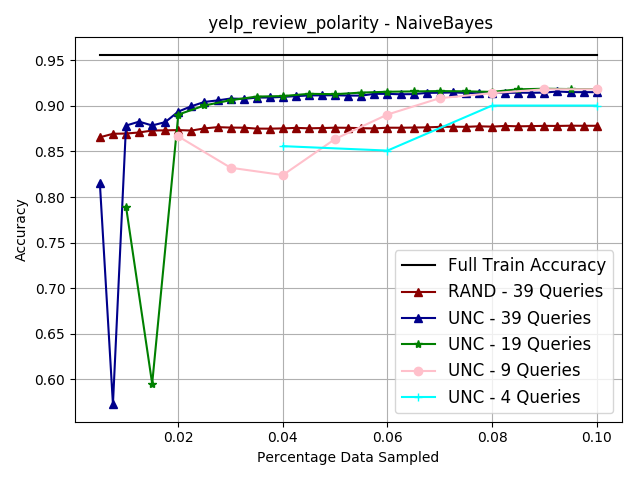}
\end{subfigure}
\begin{subfigure}[h]{0.33\linewidth}
\includegraphics[width=\linewidth]{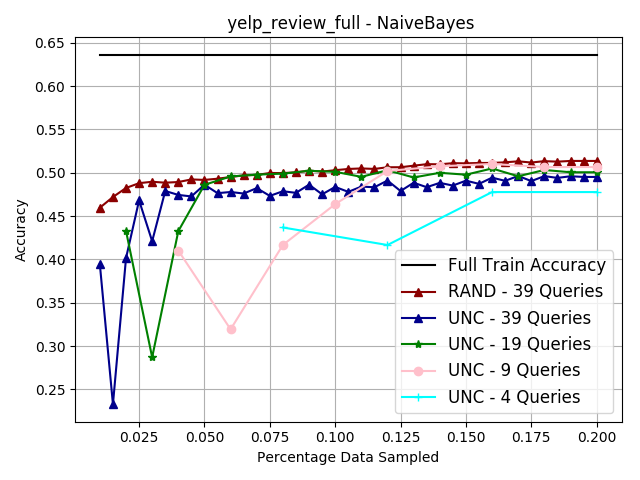}
\end{subfigure}
\begin{subfigure}[h]{0.33\linewidth}
\includegraphics[width=\linewidth]{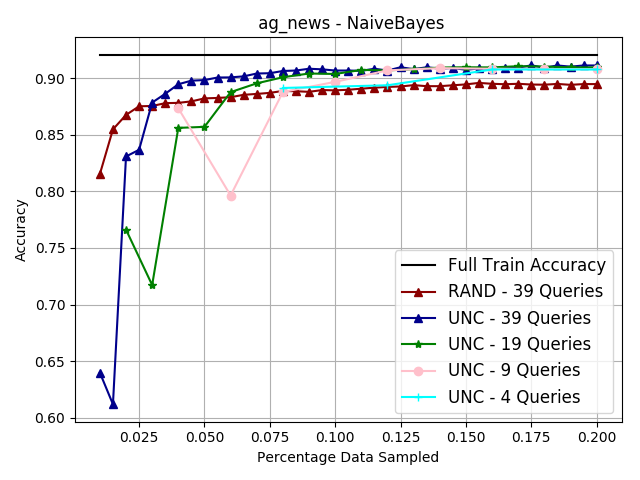}
\end{subfigure}
\caption{Accuracy across different number of queries $b$ for FastText and Naive Bayes, with  $b \times K$ constant. FastText is robust to increase in query size and significantly outperforms random in all cases} 
\label{fig:all_accuracies}
\end{figure*}

We show in Figure \ref{fig:all_accuracies} the accuracy curves for FastText and NaiveBayes, for 4, 9, 19 and 39 iterations using entropy query strategy vs random.

\end{document}